\documentclass{article} 
\usepackage{iclr2024_conference,times}

\usepackage{amsmath,amsfonts,bm}

\def\eqref#1{equation~\ref{#1}}

\def\1{\bm{1}}

\def\va{{\bm{a}}}

\def\vd{{\bm{d}}}

\def\vs{{\bm{s}}}

\def\vv{{\bm{v}}}

\DeclareMathAlphabet{\mathsfit}{\encodingdefault}{\sfdefault}{m}{sl}
\SetMathAlphabet{\mathsfit}{bold}{\encodingdefault}{\sfdefault}{bx}{n}

\providecommand{\ie}{\textit{i.e.}\@\xspace}

\usepackage{wrapfig}
\usepackage{booktabs} 
\usepackage{multirow}
\usepackage{makecell}
\usepackage{hyperref}
\usepackage{url}
\usepackage{graphicx} 
\usepackage{bbding} 
\newcommand{\tablestyle}[2]{\setlength{\tabcolsep}{#1}\renewcommand{\arraystretch}{#2}\centering\small}
\usepackage{xspace}
\newcommand{\myparagraph}[1]{{\noindent\bf #1}}
\newcommand{\methodname}{UniHSI}
\usepackage{amsmath}
\usepackage{diagbox}

\usepackage{longtable}
\usepackage[section]{placeins} 
\usepackage[misc]{ifsym} 
\newcommand\blfootnote[1]{\begingroup\renewcommand\thefootnote{}\footnote{#1}\addtocounter{footnote}{-1}\endgroup}

\title{
Unified Human-Scene Interaction via \\ Prompted Chain-of-Contacts
}

\author{
Zeqi Xiao$^{1,2}$,
Tai Wang$^{1}$,
Jingbo Wang$^{1}$,
Jinkun Cao$^{1,3}$,
Wenwei Zhang$^{1}$,
\textbf{Bo Dai}$^{1}$, \\
\ \textbf{Dahua Lin}$^{1}$,
\textbf{Jiangmiao Pang}$^{1\textrm{\Letter}}$\\
$^1$Shanghai AI Laboratory, $^2$S-Lab, NTU, $^3$CMU
}

\iclrfinalcopy 
\begin{document}

\maketitle

\blfootnote{\textrm{\Letter} Corresponding Author. Project page at this \href{https://github.com/OpenRobotLab/UniHSI}{URL}.}
\begin{figure}[h]
\vspace{-24pt}
\begin{center}
   \includegraphics[width=\linewidth]{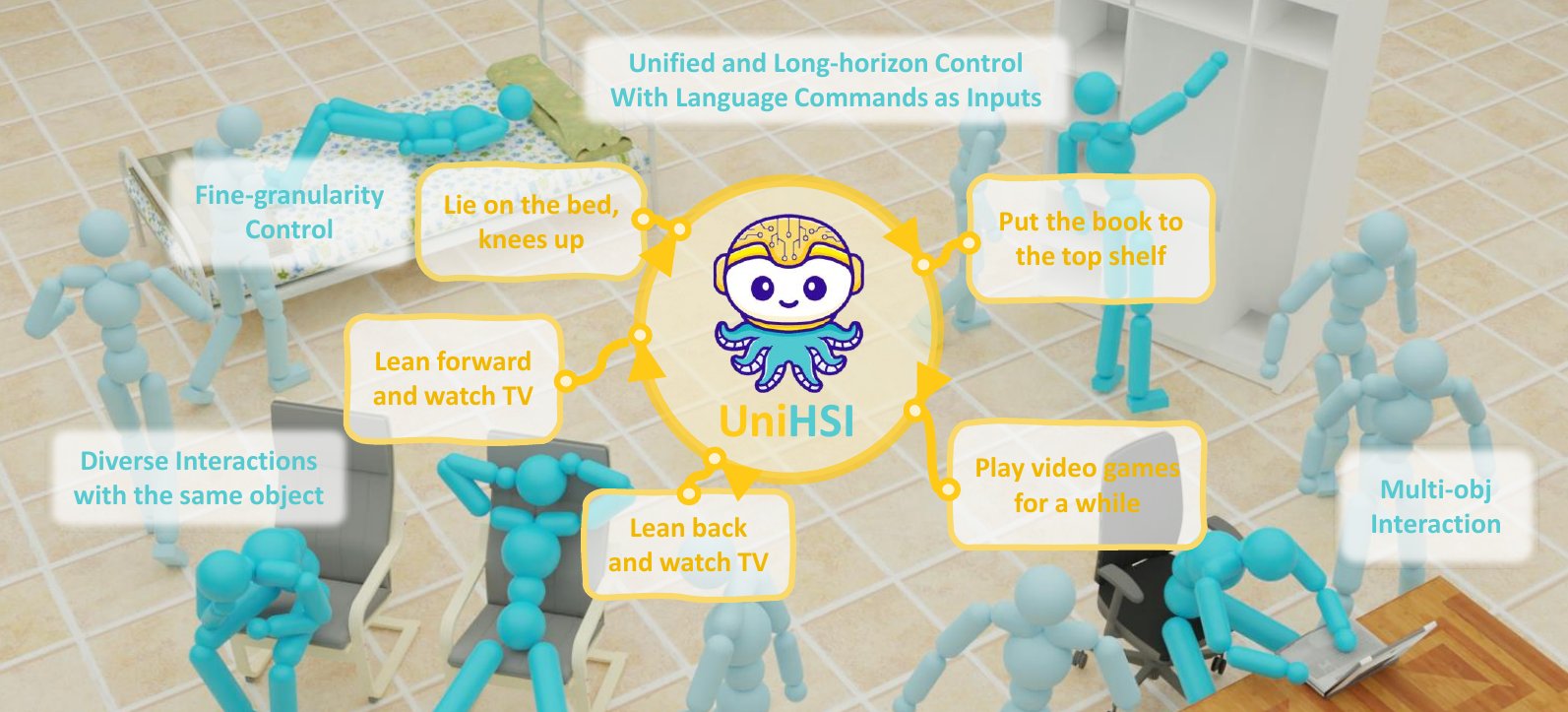}
\end{center}

    \vspace{-12pt}
   \caption{\methodname~facilitates unified and long-horizon control in response to natural language commands, offering notable features such as diverse interactions with a singular object, multi-object interactions, and fine-granularity control.}
\label{fig:teaser}
\end{figure}

\begin{abstract}
Human-Scene Interaction (HSI) is a vital component of fields like embodied AI and virtual reality. Despite advancements in motion quality and physical plausibility, 
two pivotal factors, versatile interaction control and user-friendly interfaces, require further exploration for the practical application of HSI.
This paper presents a unified HSI framework, named \emph{\methodname}, that supports unified control of diverse interactions through language commands. 
The framework defines interaction as ``Chain of Contacts (CoC)", representing steps involving human joint-object part pairs. This concept is inspired by the strong correlation between interaction types and corresponding contact regions.
Based on the definition, \methodname~constitutes a \emph{Large Language Model (LLM) Planner} to translate language prompts into task plans in the form of CoC, and a \emph{Unified Controller} that turns CoC into uniform task execution. To support training and evaluation, we collect a new dataset named \emph{ScenePlan} that encompasses thousands of task plans generated by LLMs based on diverse scenarios. Comprehensive experiments demonstrate the effectiveness of our framework in versatile task execution and generalizability to real scanned scenes.
\end{abstract}

\section{Introduction}

Human-Scene Interaction (HSI) constitutes a crucial element in various applications, including embodied AI and virtual reality. Despite the great efforts in this domain to promote motion quality \citep{holden2017phase, starke2019neural, starke2020local, hassan2021populating, zhao2022compositional, hassan2021stochastic, wang2022towards} and physical plausibility \citep{holden2017phase, starke2019neural, starke2020local, hassan2021populating, zhao2022compositional, hassan2021stochastic, wang2022towards}, two key factors, versatile interaction control and the development of a user-friendly interface, are yet to be explored before HSI can be put into practical usage.

This paper aims to provide an HSI system that supports versatile interaction control through language commands, one of the most uniform and accessible interfaces for users. Such a system requires:
1) Aligning language commands with precise interaction execution,
2) Unifying diverse interactions within a single model to ensure scalability.
To achieve this, the initial effort involves the uniform definition of different interactions. We propose that interaction itself contains a strong prior in the form of human-object contact regions. For example, in the case of ``lie down on the bed", it can be interpreted as ``first the pelvis contacting the mattress of the bed, then the head contacting the pillow". To this end, we formulate interaction as ordered sequences of human joint-object part contact pairs, which we refer to as \emph{Chain of Contacts (CoC)}.
Unlike previous contact-driven methods, which are limited to supporting specific interactions through manual design, our interaction definition is generalizable to versatile interactions and capable of modeling multi-round transitions. 
The recent advancements in Large Language Models have made it possible to translate language commands into CoC. The structured formulation then can be uniformly processed for the downstream controller to execute.

Following the above formulation, we propose \textbf{\methodname}, the first \textbf{Uni}fied physical \textbf{HSI} framework with language commands as inputs. \methodname~consists of a high-level \textbf{LLM Planner} to translate language inputs into the task plans in the form of CoC and a low-level \textbf{Unified Controller} for executing these plans. 
Combining language commands and background information such as body joint names and object part layout, we harness prompt engineering techniques to instruct LLMs to plan interaction step by step.
We design the TaskParser to support the unified execution. It serves as the core of the Unified Controller.
Following CoC, the TaskParser collects information including joint poses and object point clouds from the physical environment, then formulates them into uniform task observations and task objectives.

As illustrated in Fig. \ref{fig:teaser}, the Unified Controller models whole-body joints and arbitrary parts of objects in the scenarios to enable fine-granularity control and multi-object interaction. With different language commands, we can generate diverse interactions with the same object. Unlike previous methods that only model a limited horizon of interactions, like ``sitting down", we design the TaskParser to evaluate the completion of the current steps and sequentially fetch the next step, resulting in multi-round and long-horizon transition control. The Unified control leverages the adversarial motion prior framework \citep{peng2021amp} that uses a motion discriminator for realistic motion synthesis and a physical simulation \citep{makoviychuk2021isaac} to ensure physical plausibility.

Another impressive feature of our framework is the training is interaction annotation-free. Previous methods typically require datasets that capture both target objects and the corresponding motion sequences, which demand numerous laboring. In contrast, we leverage the interaction knowledge of LLMs to generate interaction plans. It significantly reduces the annotation requirements and makes versatile interaction training feasible. To this end, we create a novel dataset named \textbf{ScenePlan}. It encompasses thousands of interaction plans based on scenarios constructed from PartNet \citep{mo2019partnet} and ScanNet \citep{dai2017scannet} datasets. We conduct comprehensive experiments on ScenePlan. The results illustrate the effectiveness of the model in versatile interaction control and good generalizability on real scanned scenarios.

\section{Related Works}

\myparagraph{Kinematics-based Human-Scene Interaction.}
How to synthesize realistic human behavior is a long-standing topic. Most existing methods focus on promoting the quality and diversity of humanoid movements \citep{barsoum2018hp, harvey2020robust, pavllo2018quaternet, yan2019convolutional, zhang2022motiondiffuse, tevet2022human, zhang2023motiongpt} but do not consider scene influence. Recently, there has been a growing interest in synthesizing motion with human-scene interactions, driven by its applications in various applications like embodied AI and virtual reality. Many previous methods \citep{holden2017phase, starke2019neural, starke2020local, hassan2021populating, zhao2022compositional, hassan2021stochastic, wang2022towards, zhang2022couch, wang2022humanise} use data-driven kinematic models to generate static or dynamic interactions. These methods are typically inferior in physical plausibility and prone to synthesizing motions with artifacts, such as penetration, floating, and sliding. The need for additional post-processing to mitigate these artifacts hinders the real-time applicability of these frameworks.

\myparagraph{Physics-based Human-Scene Interaction.}
Recent advances in physics-based methods (e.g., \citep{peng2021amp, peng2022ase, hassan2023synthesizing, juravsky2022padl, pan2023synthesizing} hold promise for ensuring physical realism through physics-aware simulators. However, they have limitations:
1) They typically require separate policy networks for each task, limiting their ability to learn versatile interactions within a unified controller.
2) These methods often focus on basic action-based control, neglecting finer-grained interaction details.
3) They heavily rely on annotated motion sequences for human-scene interactions, which can be challenging to obtain.
In contrast, our \methodname~redesigns human-scene interactions into a uniform representation, driven by world knowledge from our high-level LLM Planner. This allows us to train a unified controller with versatile interaction skills without the need for annotated motion sequences. Key feature comparisons are in Tab. \ref{tab:feature_comparison}.

\myparagraph{Languages in Human Motion Control.}
Incorporating language understanding into human motion control has become a recent research focus. 
Existing methods primarily focus on scene-agnostic motion synthesis \citep{zhang2022motiondiffuse,chen2023executing, tevet2022motionclip,tevet2022human,zhang2023generating, zhang2023motiongpt,jiang2023motiongpt} \citep{athanasiou2023sinc}.
Generating human-scene interactions using language commands poses additional challenges because the output movements must align with the commands and be coherent with the environment. \citet{zhao2022compositional} generates static interaction gestures through rule-based mapping of language commands to specific tasks. \citet{juravsky2022padl} utilized BERT \citep{devlin2018bert} to infer language commands, but their method requires pre-defined tasks and different low-level policies for task execution. \citet{wang2022humanise} unified various tasks in a CVAE \citep{yao2022controlvae} network with a language interface, but their performance was limited due to challenges in grounding target objects and contact areas for the characters. Recently, there have been some explorations on LLM-based agent control. \citet{brohan2023rt} uses fine-tuned VLM (Vision Language Model) to directly output actions for low-level robots. \citet{rocamonde2023vision} employs CLIP-generated cos-similarity as RL training rewards.  In contrast, \methodname~utilizes large language models to transfer language commands into the formation of \emph{Chain of Contacts} and design a robust unified controller to execute versatile interaction based on the structured formation.

\begin{table*}[t]
  \vspace{-24pt}
  \footnotesize
  \begin{center}
  \caption{Comparative Analysis of Key Features between \methodname~and Preceding Methods.}
  \vspace{3pt}
  \label{tab:feature_comparison} 
  \scalebox{0.72}{\tablestyle{8pt}{1.0}
    \begin{tabular}{ccccccc}
    \toprule[1.5pt]
    Methods & \thead{Unified \\ Interaction} & \thead{Language \\ Input} & \thead{Long-horizon \\ Transition} & \thead{Interaction \\ Annotation-free} & \thead{Control \\ Joints} & \thead{Multi-object \\ Interactions} \\
    \midrule
    NSM~\cite{starke2019neural}            &            &  & \checkmark &  & 3 (pelvis, hands) & \checkmark \\
    SAMP~\cite{hassan2021stochastic}       &            & &            &  & 1 (pelvis) \\
    COUCH~\cite{zhang2022couch}            &            & &            &  & 3 (pelvis, hands) & \checkmark\\
    HUMANISE~\cite{wang2022humanise}       & \checkmark & \checkmark &            &  &
    - \\
    ScenDiffuser~\cite{huang2023diffusion} & \checkmark &  \checkmark &           &  &
    - \\
    PADL~\cite{juravsky2022padl}            &            &  \checkmark & \checkmark           &  \checkmark & - \\
    InterPhys~\cite{hassan2023synthesizing}  &            & &            & & 4 (pelvis, head, hands) \\
    \midrule
    Ours                                   & \checkmark & \checkmark & \checkmark & \checkmark    & 15 (whole-body) & \checkmark\\
    \bottomrule[1.5pt]
    \end{tabular}}
    \vspace{-18pt}
  \end{center}
\end{table*}

\section{Methodology}

\begin{figure}[t]
\begin{center}
   \includegraphics[width=\linewidth]{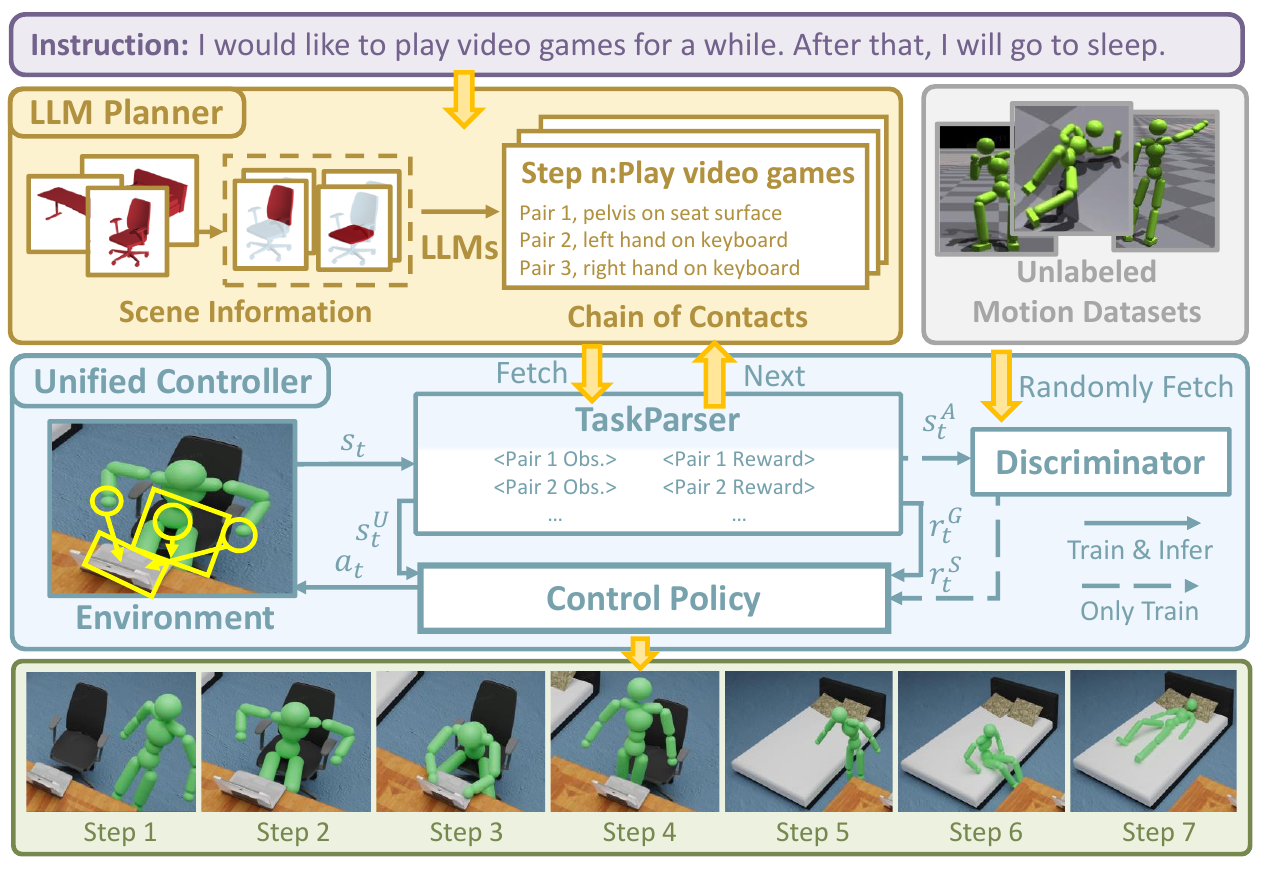}
\end{center}
\vspace{-0.5cm}
   \caption{\textbf{Comprehensive Overview of \methodname.} The entire pipeline comprises two principal components: the LLM Planner and the Unified Controller. The LLM Planner processes language inputs and background scenario information to generate multi-step plans in the form of CoC. Subsequently, the Unified Controller executes CoC step by step, producing interaction movements.}

\label{fig:main_fig}
\end{figure}


As shown in Fig. \ref{fig:main_fig}, \methodname~supports versatile human-scene interaction control following language commands. In the following subsections, we first illustrate how we design the unified interaction formulation as CoC(Sec. \ref{sec:unified_interaction_formulation}). Then we show how we translate language commands into the unified formulation by the LLM Planner (Sec. \ref{sec:llm_planner}). Finally, we elaborate on the construction of the Unified Controller (Sec. \ref{sec:unified_controller}).

\subsection{Chain of Contacts}\label{sec:unified_interaction_formulation}

The initial effort of \methodname~lies in the unified formulation of interaction. Inspired by \citet{hassan2021populating}, which infers contact regions of humans and objects based on the interaction gestures of humans, we propose a high correlation between contact regions and interaction types. Further, interactions are not limited to a single gesture but involve sequential transitions. To this end,  we can universally define interaction as CoC $\mathcal{C}$, with the formulation as
\begin{equation}
    \mathcal{C} = \{\mathcal{S}_1, \mathcal{S}_2, ...\} ,
    \label{eqn:coc}
\end{equation}
where $\mathcal{S}_i$ is the $i^{th}$ contact step. Each step $\mathcal{S}$ includes several contact pairs. For each contact pair, we control whether a joint contacts the corresponding object part and the direction of the contact. We construct each contact pair with five elements: an object $o$, an object part $p$, a humanoid joint $j$, the contact type $c$ of $j$ and $p$, and the relative direction $d$ from $j$ to $p$. The contact type includes ``contact", ``not contact", and ``not care". The relative direction includes ``up", ``down", ``front", ``back", ``left", and ``right".  For example, one contact unit $\{o, p, j, c, d\}$ could be \{chair, seat surface, pelvis, contact, up\}. In this way, we can formulate each $\mathcal{S}$ as
\begin{equation}
    \mathcal{S} = \{\{o_1, p_1, j_1, c_1, d_1\}, \{o_2, p_2, j_2, c_2, d_2\}, ...\}.
\end{equation}\label{eqn:interaction_step}

CoC is the output of the LLM Planner and the input of the Unified Controller.


\subsection{Large Language Model Planner}\label{sec:llm_planner}

\begin{figure}[t]
\begin{center}
   \includegraphics[width=\linewidth]{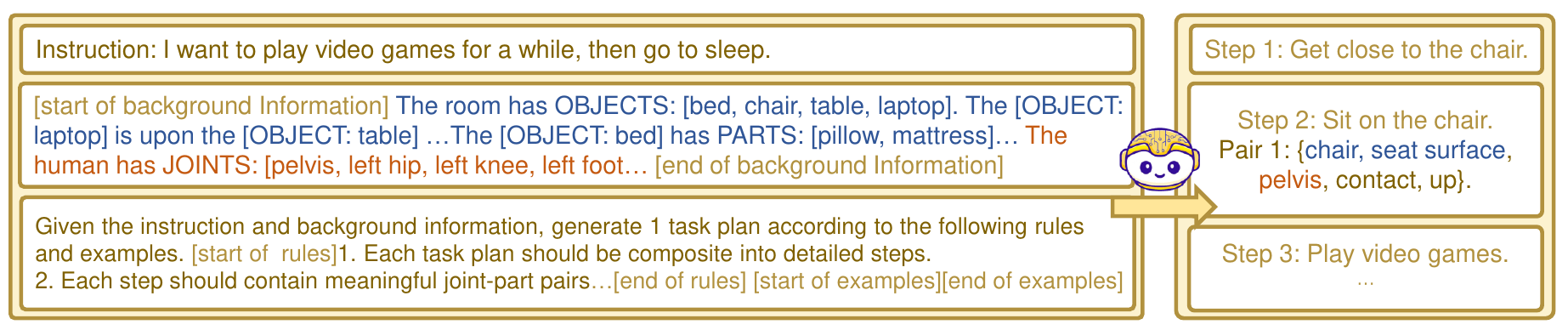}
\end{center}
\vspace{-0.5cm}
   \caption{The Procedure for Translating Language Commands into Chains of Contacts.}

\label{fig:prompt}
\end{figure}

We leverage LLMs as our planners to infer language commands $\mathcal{L}$ into manageable plans $\mathcal{C}$. 
As shown in Fig. \ref{fig:prompt}, the inputs of the LLM Planner include language commands $\mathcal{L}$, background scenario information $\mathcal{B}$, humanoid joint information $\mathcal{J}$ together with pre-set instructions, rules and examples.
Specifically, $\mathcal{B}$ includes several objects $\mathcal{O}$ and their optional spatial layouts. Each object consists of several parts $\mathcal{P}$, \ie, a chair could consist of arms, the back, and the seat. The humanoid joint information is pre-defined for all scenarios. We use prompt engineering to combine these elements together and instruct LLMs to output task plans. 
By modifying instructions in the prompts, we can generate specified numbers of plans for diverse ways of interactions. We can also let LLMs automatically generate plausible plans given the scenes. In this way, we build our interaction datasets to train and evaluate the Unified Controller.

\subsection{Unified Controller}\label{sec:unified_controller}

The Unified Controller takes multi-step plans $\mathcal{C}$ and background scenarios in the form of meshes and point clouds as input and outputs realistic movements coherent to the environments.

\myparagraph{Preliminary.} We build the controller upon AMP \citep{peng2021amp}. AMP is a goal-conditioned reinforcement learning framework incorporated with an adversarial discriminator to model the motion prior. Its objective is defined by a reward function $R(\cdot)$ as 
\begin{equation}
    R(\vs_t, \va_t, \vs_{t+1},\mathcal{G}) = w^GR^G(\vs_t, \va_t, \vs_{t+1}, \mathcal{G}) + w^SR^S(\vs_t, \vs_{t+1}).
\end{equation}
The task reward $R^G$ defines the high-level goal $\mathcal{G}$ an agent should achieve. The style reward $R^S$ encourages the agent to imitate low-level behaviors from motion datasets. $w^G$ and $w^S$ are empirical weights of $R^G$ and $R^S$, respectively. $\vs_t$, $\va_t$, $\vs_{t+1}$ are the state at time $t$, the action at time $t$, the state at time ${t+1}$, respectively. The style reward $R^S$ is modeled using an adversarial discriminator $D$, which is trained according to the objective:

\begin{equation}
\begin{aligned}
    \mathop{\mathrm{arg \ min}}_D \ -\mathbb{E}_{d^\mathcal{M}(\vs_t, \vs_{t+1})} \left[ \mathrm{log}\left(D(\vs^A_t, \vs^A_{t+1})\right) \right]
     - \mathbb{E}_{d^\pi({\vs, \vs_{t+1}})} \left[\mathrm{log}\left(1 - D(\vs^A, \vs^A_{t+1})\right) \right] \\
     + w^\mathrm{gp} \ \mathbb{E}_{d^\mathcal{M}(\vs, \vs_{t+1})} \left[\left| \left| \nabla_\phi D(\phi) \middle|_{\phi = (\vs^A, \vs^A_{t+1})} \right| \right|^2 \right],
\label{eqn:disc_loss}
\end{aligned}
\end{equation}

where $d^\mathcal{M}(\vs, \vs_{t+1})$ and $d^\pi({\vs, \vs_{t+1}})$ denote the likelihood of a state transition from $\vs_t$ to $\vs_{t+1}$ in the dataset $\mathcal{M}$ and the policy $\pi$ respectively. $w^\mathrm{gp}$ is an empirical coefficient to regularize gradient penalty. 
$\vs^A = \Phi(\vs)$ is the observation for discriminator.
The style reward $r^S = R^S(\cdot)$ for the policy is then formulated as:
\begin{equation}
    R^S(\vs_t, \vs_{t+1}) =  - \mathrm{log}(1 - D(\vs^A_t, \vs^A_{t+1})) .
\label{eqn:gan_reward}
\end{equation}

\begin{figure}[t]
\begin{center}
   \includegraphics[width=\linewidth]{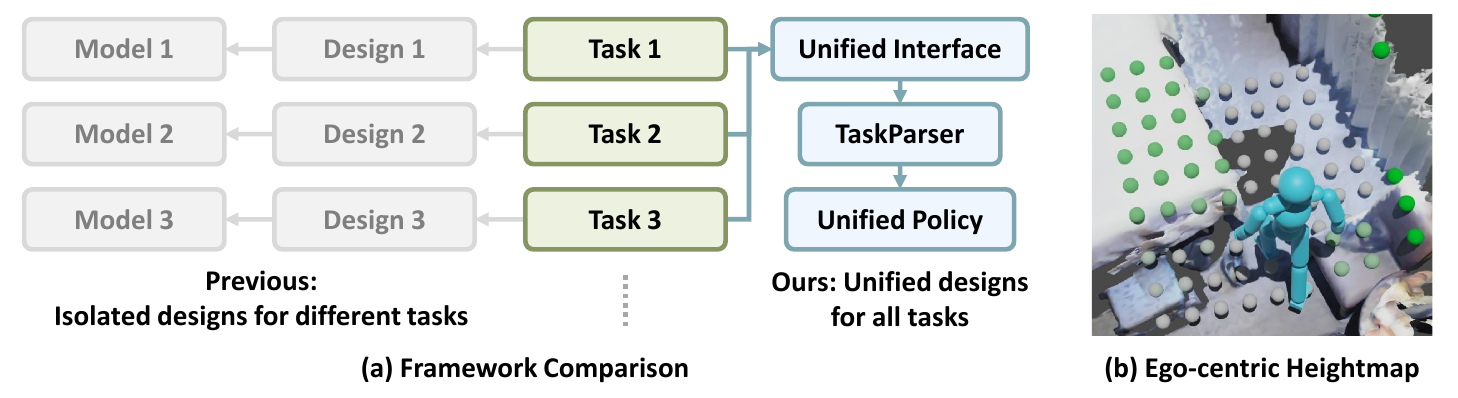}
\end{center}
\vspace{-0.5cm}
   \caption{\textbf{Design Visualization.} (a) Our framework ensures a unified design across tasks using the unified interface and the TaskParser. (b) The ego-centric height map in a ScanNet scene is depicted by green dots, with darker shades indicating greater height.}
   \vspace{-6pt}
\label{fig:design_visualization}
\end{figure}

We adopt the key design of motion discriminator for realistic motion modeling. In our implementation, we feed 10 adjacent frames together into the discriminator to assess the style. Our main contribution to the controller parts lies in unifying different tasks. As shown in the left part of Fig. \ref{fig:design_visualization} (a), AMP \citep{peng2021amp}, as well as most of the previous methods \citep{juravsky2022padl, zhao2023synthesizing}, design specified task observations, task objectives, and hyperparameters to train task-specified control policy. In contrast, we unify different tasks into Chains of Contacts and devise a TaskParser to process the uniform representation.

\myparagraph{TaskParser.}
As the core of the Unified Controller, the TaskParser is responsible for formulating CoC into uniform task observations and task objectives. It also sequentially fetches steps for multi-round interaction execution.

Given one specific contacting pair $\{o, p, j, c, d\}$, for task observation, the TaskParser collects the corresponding position $\vv^j\in\mathbb{R}^3$ of the joint $j$, and point clouds $\vv^p\in\mathbb{R}^{m\times3}$ of the object part $p$ from the simulation environment, where $m$ is the point number of point clouds. It selects the nearest point $\vv^{np}\in\vv^p$ from $\vv^p$ to $\vv^j$ as the target point for contact. We formulate task observation of the single pair as 
$\{\vv^{np}-\vv^j, c, d\}$. For the task observation in the network, we map $c$ and $d$ into digital numbers, but we still use the same notation for simplicity. Combining these contact pairs together, we get the uniform task observations $s^U = \{\{\vv^{np}_1-\vv^j_1, c_1, d_1\}, \{\vv^{np}_2-\vv^j_2, c_2, d_2\}, ..., \{\vv^{np}_n-\vv^j_n, c_n, d_n\}\}$.

The task reward $r^G=R^G(\cdot)$ is the summarization of all contact pair rewards:
\begin{equation}
    R^G = \sum_kw_kR_k,\ k=1,2,...,n.
    \label{eqn:unified_reward}
\end{equation}
We model each contact reward $R_k$ according to the contact type $c_k$. When $c_k = \mathrm{contact}$, the contact reward encourages the joint $j$ to be close to the part $p$, satisfying the specified direction $d$. When $c_k = \mathrm{not contact}$, we hope the joint $j$ is not close to the part $p$. If $c_k = \mathrm{not\ care}$, we directly set the reward to max. Following the idea, the $k^{th}$ contact reward $R_k$ is defined as 
\begin{equation}
    R_k = 
    \begin{cases}
        w_{\mathrm{dis}} \mathrm{exp}(-w_{dk}|| \vd_k ||) + w_{\mathrm{dir}} \mathrm{max}(\overline{\vd}_k\hat{\vd}_k, 0),  & c_k = \mathrm{contact} \\
        1 - \mathrm{exp}(-w_{dk}|| \vd_k ||),  & c_k = \mathrm{not\ contact} \\
        1,  & c_k = \mathrm{not\ care}\\
    \end{cases}
    \label{eqn:reward}
\end{equation}
where $\vd_k = \vv^{np}-\vv^j$ indicates the $k^{\mathrm{th}}$ distance vector, $\overline{\vd}_k$ is the normalized unit vector of $\vd_k$, $\hat{\vd}_k$ is the unit direction vector specified by direction $d_k$, and $c_k$ is the $k^{\mathrm{th}}$ contact type. $w_{dis}$, $w_{dir}$, $w_{dk}$ are corresponding weights. We set the scale interval of $R_k$ as $[0, 1]$ and use \emph{exp} to ensure it.

Similar to the formulation of contact reward, the TaskParser considers a step to be completed if All $k=1,2,...,n$ satisfy: if $c_k = \mathrm{contact}: || \vd_k ||<0.1\ \mathrm{and}\ \overline{\vd}_k\hat{\vd}_k>0.8$, if $c_k = \mathrm{not\ contact}: || \vd_k ||>0.1$, if $c_k = \mathrm{not\ care}, True$.

\myparagraph{Adaptive Contact Weights.} The formulation of \ref{eqn:unified_reward} includes lots of weights to balance different contact parts of the rewards. Empirically setting them requires much laboring and is not generalizable to versatile tasks. To this end, we adaptively set these weights based on the current optimization process. The basic idea is to give parts of rewards that are hard to optimize high rewards while lowering the weights of easier parts. Given $R_1$, $R_2$, ...,  $R_n$, we heuristically set their weights to
\begin{equation}
    w_k = \frac{1-R_k}{n-\sum_{k=1,2,...,n}R_k+e},
\end{equation}
    
\myparagraph{Ego-centric Heightmap.}
The humanoid must be scene-aware to avoid collision when navigating or interacting in a scene. We adopt similar approaches in \citet{wang2022towards, won2022physics, starke2019neural} that sample surrounding information as the humanoid's observation. We build1 a square ego-centric heightmap that samples the height of surrounding objects (Fig. \ref{fig:design_visualization} (b)). It is important to extend our methods into real scanned scenarios such as ScanNet \citep{dai2017scannet} in which various objects are densely distributed and easily collide.

\begin{table*}[t]
  \footnotesize
  \begin{center}
  \caption{Performance Evaluation on the ScenePlan Dataset.}
  \vspace{-6pt}
  \label{tab:performance_on_sceneplan} 
  \scalebox{0.9}{\tablestyle{8pt}{1.0}
    \begin{tabular}{c|c|c|c|c|c|c|c|c|c}
    \toprule[1.5pt]
    \multirow{2}*{Source}
    & \multicolumn{3}{c|}{Success Rate (\%) $\uparrow$}
    & \multicolumn{3}{c|}{Contact Error $\downarrow$}
    & \multicolumn{3}{c}{Success Steps}\\
    & Simple & Mid & Hard & Simple & Mid & Hard & Simple & Mid & Hard\\
    \midrule
    PartNet \citep{mo2019partnet} & 85.5 & 67.9   & 40.5 & 0.035 & 0.037 & 0.040 & 2.1 & 4.1 & 4.8\\
    wo Adaptive Weights  & 21.2 & 5.3 & 0.1 & 0.181 & 0.312 & 0.487 & 0.7 & 1.2 & 0.0 \\
    wo Heightmap  & 61.6 & 45.7 & 0.0  & 0.068 & 0.076 & - & 1.8 & 3.4 & 0.0 \\
    \midrule
    ScanNet \citep{dai2017scannet} & 73.2 & 43.1 & 22.3 & 0.061 & 0.072 & 0.062 & 2.2 & 3.5 & 4.8   \\
    \bottomrule[1.5pt]
    \end{tabular}}
    
\vspace{-18pt}
  \end{center}
\end{table*}

\section{Experiments}
Existing methods and datasets related to human-scene interactions mainly focus on short and limited tasks \citep{hassan2021stochastic, peng2021amp, hassan2023synthesizing, wang2022humanise}. To the best of our knowledge, we are the first method that supports arbitrary horizon interactions with language commands as input. To this end, we construct a novel dataset for training and evaluation. We also conduct various ablations with vanilla baselines and key components of our framework.

\begin{figure}[t]
\begin{center}
   \includegraphics[width=\linewidth]{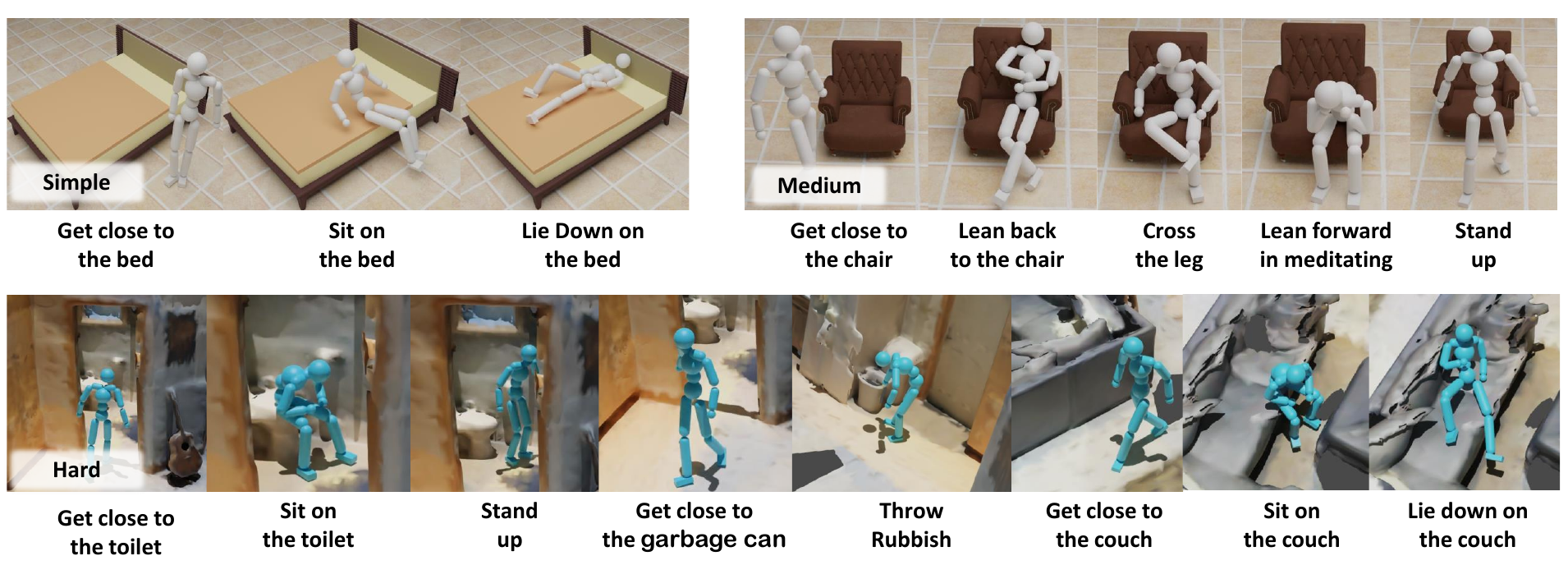}
\end{center}
\vspace{-0.5cm}
   \caption{Visual Examples Illustrating Tasks of Varying Difficulty Levels.}
\label{fig:task_examples}
\vspace{-12pt}
\end{figure}

\subsection{Datasets and Metrics}

To facilitate the training and evaluation of \methodname, we construct a novel ScenePlan dataset comprising various indoor scenarios and interaction plans.
The indoor scenarios are collected and constructed from object datasets and scanned scene datasets. We leverage our LLM Planner to generate interaction plans based on these scenarios.
The training of our model also requires motion datasets to train the motion discriminator, which constrains our agents to interact in natural ways.
We follow the practice of \citet{hassan2023synthesizing} to evaluate the performance of our method.

\myparagraph{ScenePlan.}\label{sec:scene_plan}
We gather scenarios for ScenePlan from PartNet \citep{mo2019partnet} and ScanNet \citep{dai2017scannet} datasets. PartNet offers indoor objects with fine-grained part annotations, ideal for LLM Planners. We select diverse objects from PartNet and compose them into scenarios.
For ScanNet, which contains real indoor room scenes, we collect scenes and annotate key object parts based on fragmented area annotations. We then employ the LLM Planner to generate various interaction plans from these scenarios.
Our training set includes 40 objects from PartNet, with 5-20 plausible interaction steps generated for each object. During training, we randomly choose 1-4 objects from this set for each scenario and select their steps as interaction plans.
The evaluation set consists of 40 PartNet objects and 10 ScanNet scenarios. We construct objects from PartNet into scenarios either manually or randomly. We generated 1,040 interaction plans for PartNet scenarios and 100 interaction plans for ScanNet scenarios. These plans encompass diverse interactions, including different types, horizons, and multiple objects.

\myparagraph{Motion Datasets.}
We use the SAMP dataset \citep{hassan2021stochastic} and CIRCLE \citep{araujo2023circle} as our motion dataset. SAMP includes 100 minutes of MoCap clips, covering common walking, sitting, and lying down behaviors. CIRCLE contains diverse right and left-hand reaching data. We use all clips in SAMP and pick 20 representative clips in CIRCLE for training.

\myparagraph{Metrics.}
We follow \citet{hassan2023synthesizing} that uses \emph{Success Rate} and \emph{Contact Error} (\emph{Precision} in \citet{hassan2023synthesizing}) as the main metrics to measure the quality of interactions quantitatively. Success Rate records the percentage of trials that humanoids successfully complete every step of the whole plan. In our experiments, we consider a trial of $n$ steps to be successfully completed if humanoids finish it in $n\times10$ seconds. We also record the average error of all contact pairs:
\begin{equation}
    \mathrm{Contact Error} = \sum_{i,c_i\neq 0}er_i/\sum_{i,c_i\neq 0}1, \qquad
    er_i = 
    \begin{cases}
        || \vd_k ||,  & c_i = \mathrm{contact} \\
        \mathrm{min}(0.3-|| \vd_k ||,0).  & c_i = \mathrm{not\ contact}
    \end{cases}
    \label{eqn:rewardSit}
\end{equation}
We further record \emph{Success Steps}, which denotes the average success step in task execution.

\subsection{Performance on ScenePlan}
We initially conducted experiments on our ScenePlan dataset. To measure performance in detail, we categorize task plans into three levels: simple, medium, and hard. We classify plans within 3 steps as simple tasks, those with more than 3 steps but with a single object as medium-level tasks, and those with multiple objects as hard tasks. Simple task plans typically involve straightforward interactions. Medium-level plans encompass more diverse interactions with multiple rounds of transitions. Hard task plans introduce multiple objects, requiring agents to navigate between these objects and interact with one or more objects simultaneously. Examples of tasks are illustrated in Fig. \ref{fig:task_examples}. 

As shown in Table \ref{tab:performance_on_sceneplan}, \methodname~performs well in simple task plans, exhibiting a high Success Rate and low Error. However, as task plans become more diverse and complex, the performance of our model experiences a noticeable decline. Nevertheless, the Success Steps metric continues to increase, indicating that our model still performs well in parts of the plans. It's important to note that the scenarios in the ScenePlan test set are unseen during training, and scenes from ScanNet exhibit a modality gap with the training set. The overall performance on the test set demonstrates the versatile capability, robustness, and generalization ability of \methodname.

\begin{table*}[t]
  \footnotesize
  \begin{center}
  \vspace{-36pt}
  \caption{Ablation Study on Baseline Models and Vanilla Implementations.}
  \vspace{3pt}
  \label{tab:ablation} 
  \scalebox{0.90}{\tablestyle{8pt}{1.0}
    \begin{tabular}{l|c|c|c|c|c|c}
    \toprule[1.5pt]
    \multirow{2}*{Methods}
    & \multicolumn{3}{c|}{Success Rate (\%) $\uparrow$}
    & \multicolumn{3}{c}{Contact Error $\downarrow$}\\
    & Sit & Lie Down & Reach & Sit & Lie Down & Reach \\
    \midrule
    NSM  - Sit \citep{starke2019neural}                   & 75.0 & -    & - & 0.19 & -    & - \\
    SAMP - Sit \citep{hassan2021stochastic}        & 75.0 & - & - & 0.06 & - & - \\
    SAMP - Lie Down\citep{hassan2021stochastic}        & - & 50.0 & - & - & 0.05 & - \\
    InterPhys - Sit \citep{hassan2023synthesizing} & 93.7 & - & - & 0.09 & - & - \\
    InterPhys - Lie Down\citep{hassan2023synthesizing} & - & 80.0 & - & - & 0.30 & - \\
    \midrule
    AMP \citep{peng2021amp}-Sit                            & 77.3 & -    & -    & 0.090 & -     & -     \\
    AMP-Lie Down                       & -    & 21.3    & - & - & 0.112     & -     \\
    AMP-Reach                       & -    & -   & \textbf{98.1} & - & -    & 0.016     \\
    AMP-Vanilla Combination (VC)       & 62.5    & 20.1   & 90.3 & 0.093 & 0.108    & 0.032     \\
    \midrule
    \methodname                             & \textbf{94.3} & \textbf{81.5} & 97.5 & \textbf{0.032} & \textbf{0.061} & \textbf{0.016} \\
    \bottomrule[1.5pt]
    \end{tabular}}
  \end{center}
\end{table*}
\begin{figure}[t]
\vspace{-12pt}
\begin{center}
   \includegraphics[width=\linewidth]{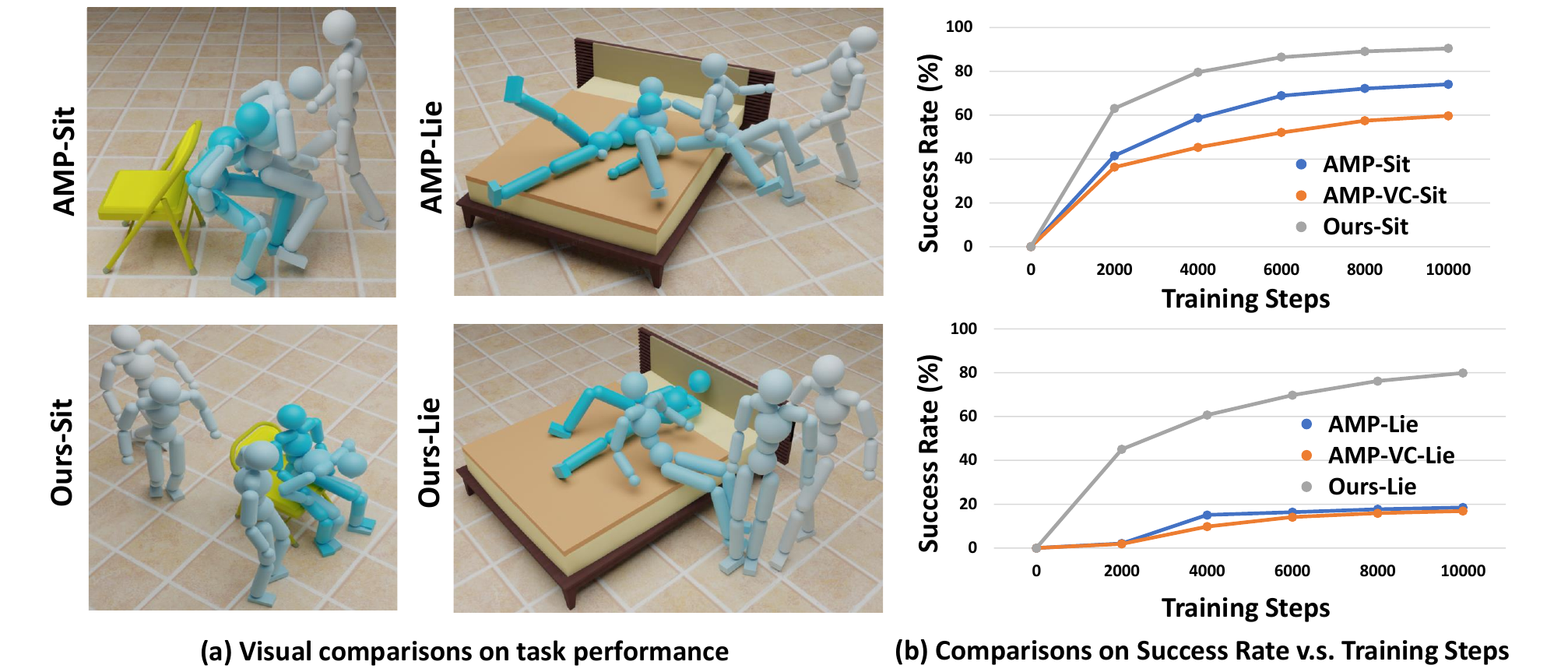}
\end{center}
\vspace{-12pt}
   \caption{\textbf{Visual Ablations.} (a) Our model exhibits superior natural and accurate performance compared to baselines in tasks such as ``Sit" and ``Lie Down". (b) Our model demonstrates more efficient and effective training procedures.}
\vspace{-18pt}
\label{fig:ablation}
\end{figure}

\vspace{-12pt}
\subsection{Ablation Studies}

\subsubsection{Key Components Ablation}\label{ablation:key_components}

\myparagraph{Choice of LLMs for UniHSI.} We evaluated different Language Model (LM) choices
\begin{wraptable}{r}{0.40\textwidth}\label{LLM_ablation}
\small
	\centering
        \vspace{-18pt}
         \caption{UniHSI with different LLMs.}
         \vspace{3pt}
	\begin{tabular}{l|c|c}
        \toprule[1.5pt]
        LLM Type & ESR (\%) $\uparrow$ & PC (\%) $\uparrow$\\
        \midrule[1pt]
        Human & 73.2 & - \\
	w. GPT-3.5 & 35.6 & 49.1 \\
        w. GPT-4 & 57.3 & 71.9 \\
        \bottomrule[1.5pt]
	\end{tabular}
        \vspace{-12pt}
\end{wraptable}
for the LLM Planner using 100 sets of language commands. We compared task plan Execution Success Rate (ESR) and Planning Correctness (PC) among humans, GPT-3.5\cite{gpt3}, and GPT-4\cite{openai2023gpt4} across 10 tests per plan. PC is evaluated by humans, with choices of "correct" and "not correct". GPT-4 outperformed GPT-3.5, but both LLMs still lag behind human performance. Failures typically involved incomplete planning and out-of-distribution interactions, like GPT-3.5 occasionally skipping transitions or generating out-of-distribution actions like opening a laptop. While using more rules in prompts and GPT-4 can mitigate these issues, errors can still occur.

\myparagraph{Adaptive Weights.} Table \ref{tab:performance_on_sceneplan} demonstrates that removing Adaptive Weights from our controller leads to a substantial performance decline across all task levels. Adaptive Weights are crucial for optimizing various contact pairs effectively. They automatically adjust weights, reducing them for unused or easily learned pairs and increasing them for more challenging pairs. This becomes especially vital as tasks become more complex.

\myparagraph{Ego-centric Heightmap.} Removing the Ego-centric Heightmap results in performance degradation, especially for difficult tasks. This heightmap is essential for agent navigation within scenes, enabling perception of surroundings and preventing collisions with objects. This is particularly critical for challenging tasks involving complex scenarios and numerous objects. Additionally, the Ego-centric Heightmap is key to our model's ability to generalize to real scanned scenes.

\vspace{-6pt}
\subsubsection{Design Comparison with Previous Methods}\label{ablation:compare_baseline}
\vspace{-6pt}

\myparagraph{Baseline Settings.} 
We compared our approach to previous methods using simple interaction tasks like ``Sit," ``Lie Down," and ``Reach." Direct comparisons are challenging due to differences in training data and code unavailability for a closely related method \citep{hassan2023synthesizing, starke2019neural, hassan2021stochastic}. Thus we list the results from their papers and implement a simple version of InterPhys \citep{hassan2023synthesizing}. We integrated key design elements from \citet{hassan2023synthesizing} into our baseline model \citep{peng2021amp} to ensure fairness. Task observations and objectives were manually formulated for various tasks, following \citet{hassan2023synthesizing}, with task objectives expressed as:

\begin{equation}
R^G =
\begin{cases}
0.7 R^\mathrm{near} + 0.3 R^\mathrm{far}, & \text{if distance} > 0.5\text{m} \\
0.7 R^\mathrm{near} + 0.3, & \text{otherwise}\\
\end{cases}
\label{eqn:baseline_objective}
\end{equation}

In this equation, $R^\mathrm{far}$ encourages character movement toward the object, and $R^\mathrm{near}$ encourages specific task performance when the character is close, necessitating task-specific designs.

We also created a vanilla baseline by consolidating multiple tasks within a single model. We combined task observations from various tasks and included task choices within these observations. We randomly selected tasks and trained them with their respective rewards during training. This experiment involved a total of 70 objects (30 for sitting, 30 for lying down, and 10 for reaching) with 4096 trials per task and random variations in orientation and object placement during evaluation.

\myparagraph{Quantitative Comparison.}
In Table \ref{tab:ablation}, \methodname~consistently outperforms or matches baseline implementations across various metrics. The performance advantage is most pronounced in complex tasks, especially the challenging ``Lie Down" task. This improvement stems from our approach of breaking tasks into multi-step plans, reducing task complexity. Additionally, our model benefits from shared motion transitions among tasks, enhancing its adaptability.
Figure \ref{fig:ablation} (b) shows that our methods achieve higher success rates and converge faster than baseline implementations. Importantly, the vanilla combination of AMP \citep{peng2021amp} results in a noticeable performance drop in all tasks while our methods remain effective. This difference is because the vanilla combination introduces interference and inefficiencies in training, whereas our approach unifies tasks into consistent representations and objectives, enhancing multi-task learning.

\myparagraph{Qualitative Comparison.} In Figure \ref{fig:ablation} (a), we qualitatively visualize the performance of baseline methods and our model. Our model performs more naturally and accurately than the baselines in tasks like ``Sit" and ``Lie Down". This is primarily attributed to the differences in task objectives. Baseline objectives (Eq. \ref{eqn:baseline_objective}) model the combination of sub-tasks, such as walking close and sitting down, as simultaneous processes. Consequently, agents tend to perform these different goals simultaneously. For example, they may attempt to sit down even if they are not in the correct position or throw themselves like a projectile onto the bed, disregarding the natural task progression. On the other hand, our methods decompose tasks into natural movements through language planners, resulting in more realistic interactions.

\vspace{-12pt}

\section{Conclusion}
\vspace{-12pt}
UniHSI is a unified Human-Scene Interaction (HSI) system adept at diverse interactions and language commands. Defined as Chains of Contacts (CoC), interactions involve sequences of human joint-object part contact pairs. UniHSI integrates a Large Language Planner for command translation into CoC and a Unified Controller for uniform execution. Comprehensive experiments showcase UniHSI's effectiveness and generalizability, representing a significant advancement in versatile and user-friendly HSI systems.
\myparagraph{Acknowledgement.} We acknowledge Shanghai AI Lab and NTU S-Lab for their funding support.

\appendix
\section{Limitations and Future Work.}
Apart from the advantages of our framework, there are a few limitations. First, our framework can only control humanoids to interact with fixed objects. We do not take moving or carrying objects into consideration. Enabling humanoids to interact with movable objects is an important future direction. Besides, we do not integrate LLM seamlessly into the training process. In the current design, we use pre-generated plans. Involving LLM in the training pipeline will promote the scalability of interaction types and make the whole framework more integrated.

\section{Implementation Details}
We follow \citet{peng2021amp} to construct the low-level controller, including a policy and discriminator networks. The policy network comprises a critic network and an actor network, both of which are modeled as a CNN layer followed by two MLP layers with [1024, 1024, 512] units. The discriminator is modeled with two MLP layers having [1024, 1024, 512] units. We use PPO \citep{schulman2017proximal} as the base reinforcement learning algorithm for policy training and employ the Adam optimizer \citet{kingma2014adam} with a learning rate of 2e-5. Our experiments are conducted on the IsaacGym \citep{makoviychuk2021isaac} simulator using a single Nvidia A100 GPU with 8192 parallel environments.

\section{Detailed prompting example of the LLM Planner}
As shown in Table. \ref{tab:detailed_prompt_example}. We present the full prompting example of the input and output of the LLM Planner that is demonstrated in Fig. 2 and Fig. 3 of the main paper. The output is generated by \cite{gpt3}. Notably, in Tab. \ref{tab:detailed_prompt_example}, example 1 step 2 pair 2: the OBJECT is the chair and PART is the left knee. It's a design choice. Our framework supports interactions between joints. We model the interaction between joints in the same way as the interaction with objects. We only need to replace the point cloud of the object part with a joint position. Some parts of the plans involve "walking to a specific place," which do not contain contacts. To model these special cases in our representations and execute them uniformly, we treat them as a pseudo contact: contacting the pelvis (root) to the target place point. This allows the policy to output a "walking" movement. We represent such cases as \{object, none, none, none, direction\}. In the future study, we will collect a list of language commands and integrate ChatGPT \cite{gpt3} and GPT \cite{openai2023gpt4} into the loop to evaluate the performance of the whole framework of \methodname.

\begin{figure}[t]
\begin{center}
   \includegraphics[width=\linewidth]{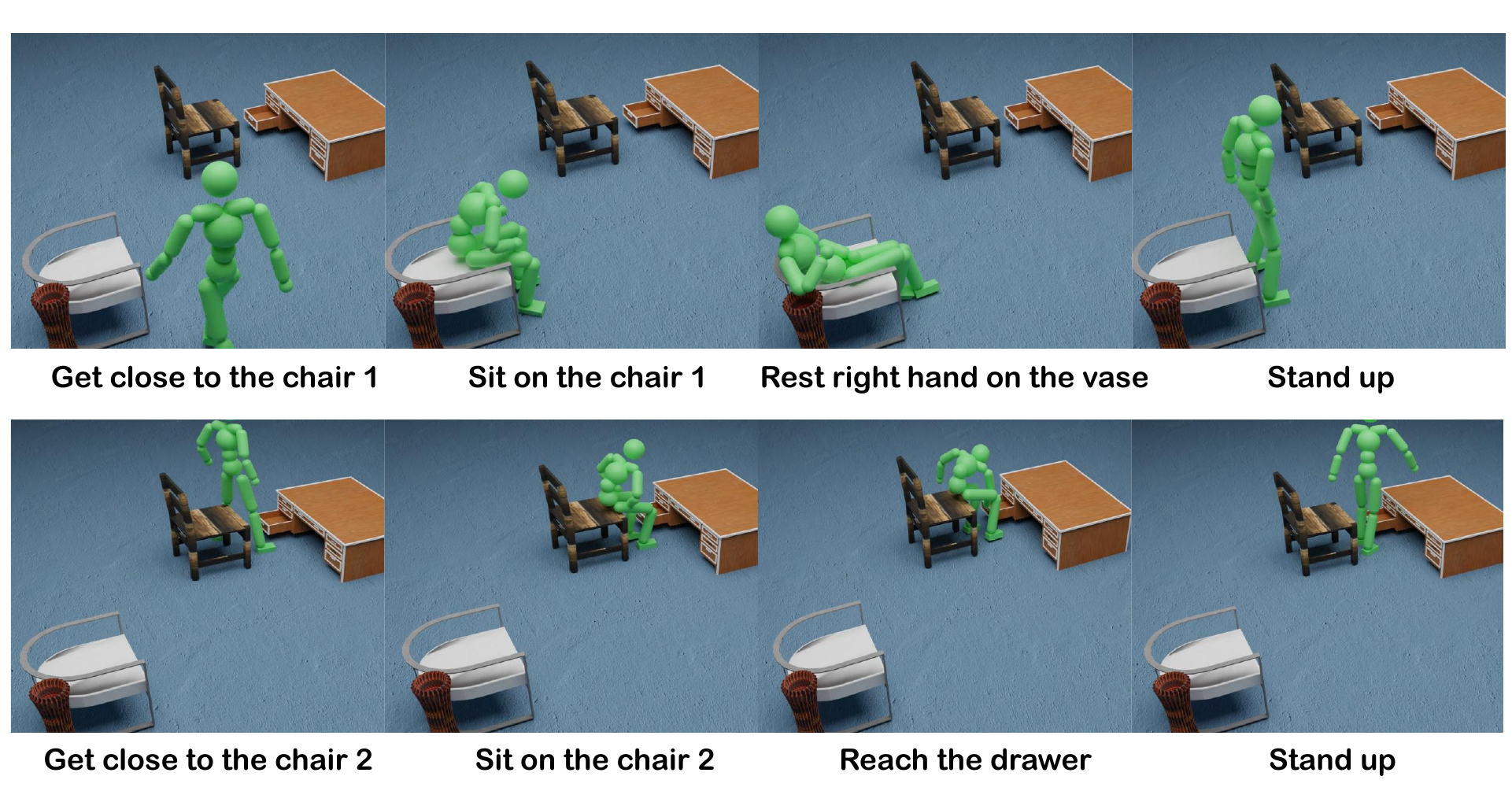}
\end{center}
\vspace{-0.5cm}
   \caption{Illustration of a Multi-Object Interaction Scenario.}
\label{fig:multi_obj_visual}
\end{figure}

\begin{figure}[t]
\begin{center}
   \includegraphics[width=\linewidth]{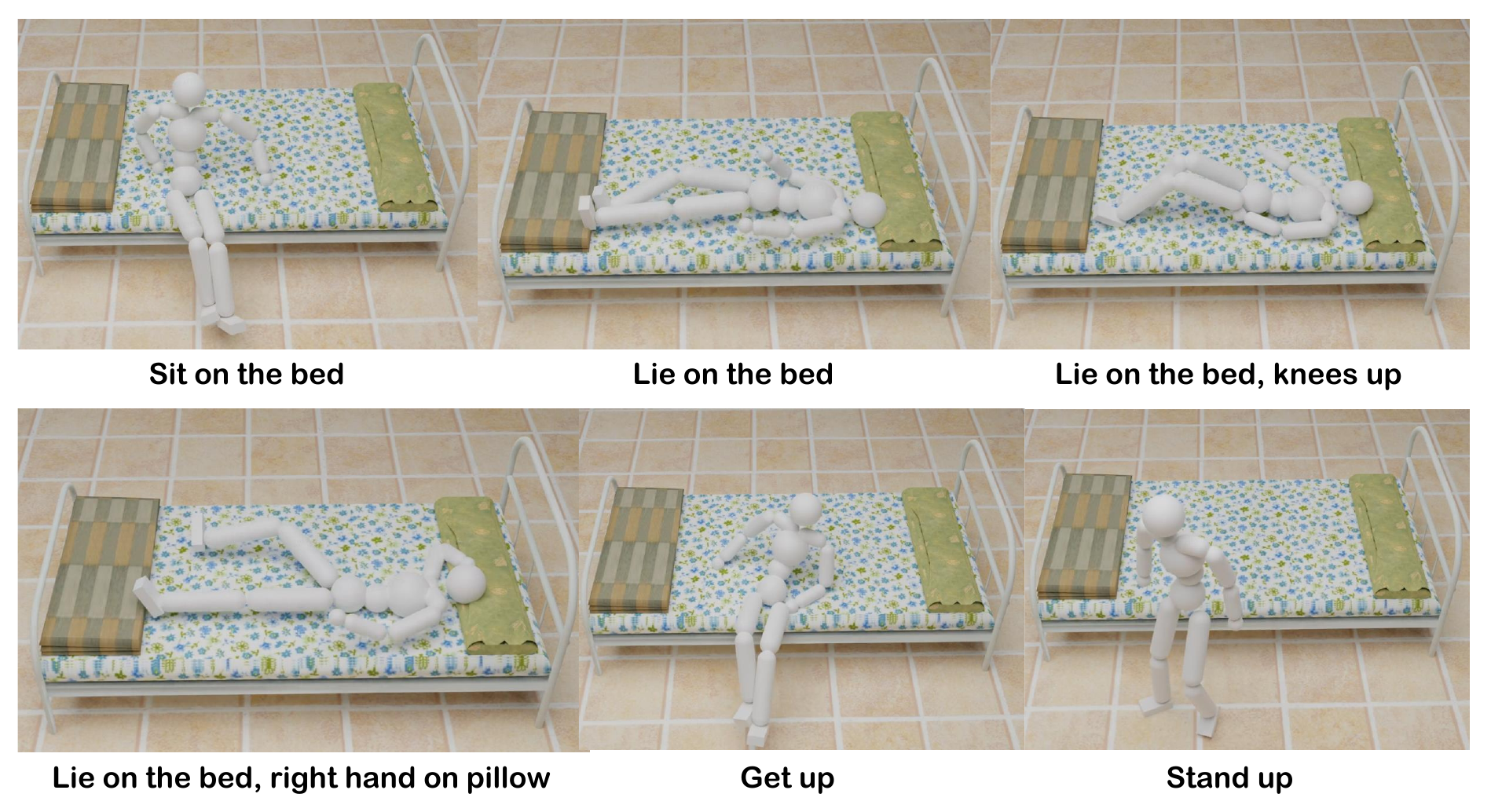}
\end{center}
\vspace{-0.5cm}
   \caption{Illustration of a Multi-Step Interaction Involving the Same Object.}
\label{fig:multi_step_visual}

\end{figure}

\section{Details of the ScenePlan}
We present three examples of different levels of interaction plans in the ScenePlan in Table \ref{tab:simple_example}, \ref{tab:mid_example}, and \ref{tab:long_example}, respectively. Simple-level interaction plans involve interactions within 3 steps and with 1 object. Medium-level interaction
plans involve more than 3 steps with 1 object. Hard-level interaction plans
involve interactions of more than 3 steps and more than 1 object. Specifically, each interaction plan has an item number and two subitems named "obj" and "chain\_of\_contacts". The "obj" item includes information about objects like object ID, name, and transformation parameters. The "chain\_of\_contacts" item includes steps of contact pairs in the form of CoC.

We provide the list of interaction types that are included in the training and evaluation of our framework in Table \ref{tab:sceneplan_list_1} and \ref{tab:sceneplan_list_2}.

\section{More Visualizations}
\begin{figure}[t]
\begin{center}
   \includegraphics[width=\linewidth]{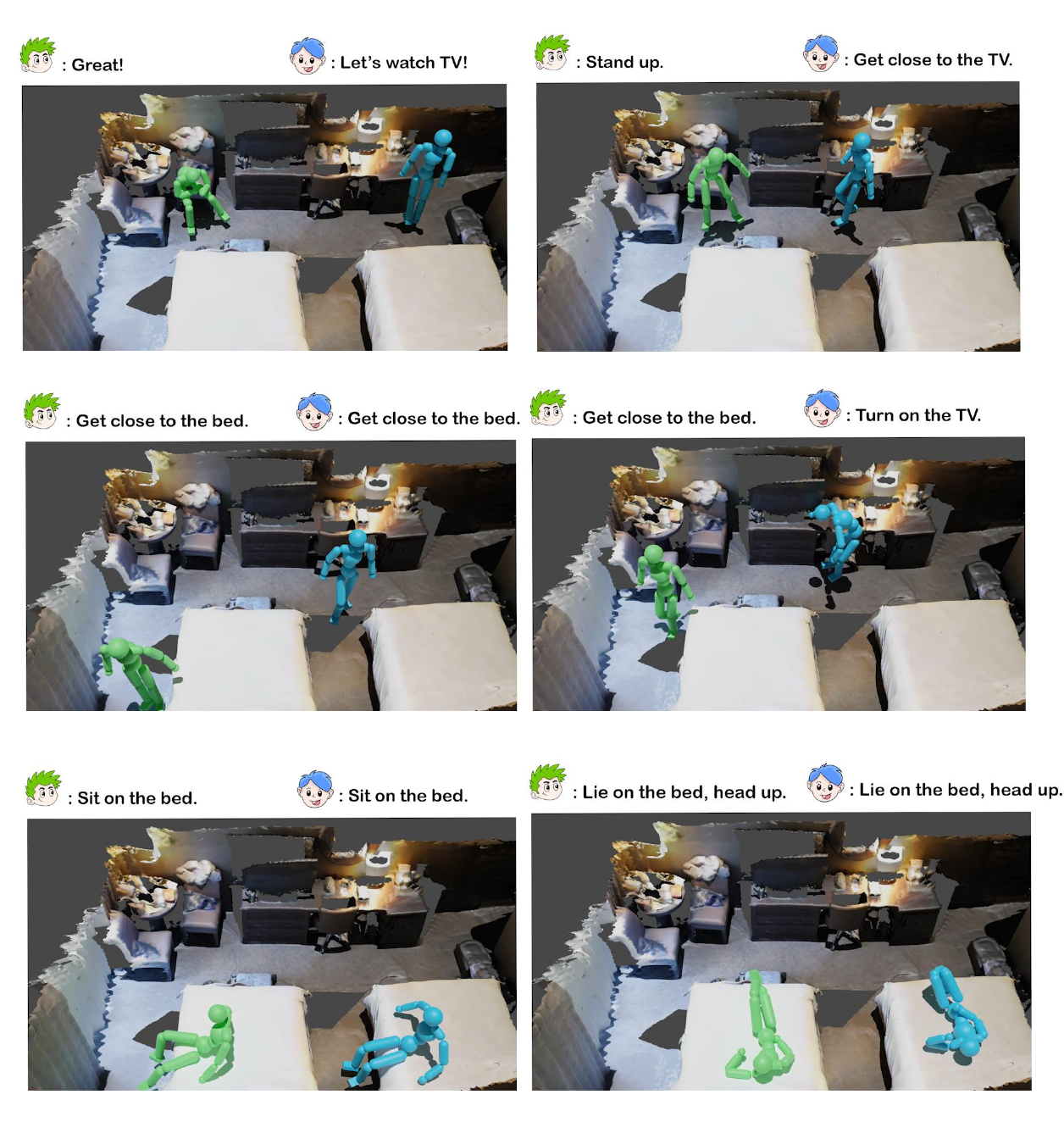}
\end{center}
\vspace{-0.5cm}
   \caption{\textbf{Illustration of Multi-Agent Interaction.} Note: Achieving ``multi-agent interaction" is presently limited to the commands level.}
\label{fig:multi_agent_visual}
\end{figure}

We further provide more quantitative results in Fig. \ref{fig:multi_obj_visual}, \ref{fig:multi_step_visual}, \ref{fig:multi_agent_visual}.

\section{Demonstration of failure planning}
In Table \ref{failure_planning}, we showcase instances where LLMs encounter challenges in generating accurate plans.
We bolded the failure in the plans.
Plans produced by LLMs may occasionally falter in spatial relations. This issue is mainly attributed to their suboptimal grasp of spatial contexts. Furthermore, LLMs may occasionally devise plans involving object manipulation, presenting difficulties in successful execution at the current stage.

\begin{table*}[t]
  \footnotesize
  \begin{center}
  \vspace{-12pt}
  \caption{Demonstration of Failure Planning.}
  \vspace{3pt}
  \scalebox{0.90}{\tablestyle{8pt}{1.0}
  \label{failure_planning}
    \begin{tabular}{p{4.5cm}|p{4.5cm}|p{4.5cm}}
    \toprule[1.5pt]
    \multicolumn{3}{l}{Instruction: I want to get up from the bed and watch movies.} \\
    \midrule
    \multicolumn{3}{l}{Output}  \\
    \midrule
    Human & GPT-3.5 \citep{gpt3} & GPT-4 \citep{openai2023gpt4} \\
    \midrule
\makecell[l]{Step 1: Get up from the bed.\\
Pair 1: \{bed, mattress, torso, not \\contact, up\}\\
Pair 2: \{bed, mattress, pelvis, contact, \\up\}\\
Step 2: Stand up from the bed.\\
Pair 1: \{bed, none, none, none, front\}\\
Step 3: Get close to the chair.\\
Pair 1: \{chair, none, none, none, \\front\}\\
Step 4: Sit on the chair.\\
Pair 1: \{chair, seat surface, pelvis, \\contact, up\}\\
Step 5: control the laptop.\\
Pair 1: \{laptop, keyboard, right hand, \\contact, up\}\\
Pair 2: \{laptop, keyboard, left hand, \\contact, up\}\\} 
&
\makecell[l]{Step 1: Get up from the bed.\\
Pair 1: \{bed, none, none, none, \textbf{up}\}\\
Step 2: Get close to the chair.\\
Pair 1: \{chair, none, none, none, \\front\}\\
Step 3: \textbf{Open the laptop.} \\
Pair 1: \{laptop, screen, left hand, \\contact, up\}\\
Pair 2: \{laptop, screen, right hand, \\contact, up\}\\
Step 4: Search for movies on the \\laptop.\\
Pair 1: \{laptop, keyboard, left hand, \\contact, up\}\\
Pair 2: \{laptop, keyboard, right hand, \\contact, up\}\\} 
&
\makecell[l]{Step 1: Get up from the bed.\\
Pair 1: \{bed, mattress, torso, not \\contact, up\}\\
Step 2: Move away from the bed.\\
Pair 1: \{bed, none, none, none, \textbf{back}\}\\
Step 3: Get close to the chair.\\
Pair 1: \{chair, none, none, none, \\front\}\\
Step 4: Sit on the chair.\\
Pair 1: \{chair, seat surface, pelvis, \\contact, up\}\\
Step 5: Use the keyboard of the \\laptop to start a movie.\\
Pair 1: \{laptop, keyboard, right hand, \\contact, up\}\\
Pair 2: \{laptop, keyboard, left hand, \\contact, up\}\\ } \\
\bottomrule[1.5pt]
    \end{tabular}}
  \end{center}
\end{table*}

\section{User Study on Motion Reality.}
To examine the global reality of the generated motion, we further conducted a user study on the evaluation of motion reality. The results are presented in the Table \ref{tab:user_study}. The Naturalness score, ranging from 0 to 5, reflects the degree of perceived naturalness, with higher scores indicating a more natural movements. Similarly, the Semantic Faithfulness score ranges from 0 to 5. A higher score denotes a greater alignment with the semantic input.

However, quantitative evaluation is challenging at this stage and requires further exploration.

\begin{table*}[h!]
  \footnotesize
  \begin{center}
  \vspace{-12pt}
  \caption{User Study on Motion Reality.}
  \vspace{3pt}
  \label{tab:user_study} 
  \scalebox{0.90}{\tablestyle{8pt}{1.0}
    \begin{tabular}{p{4.5cm}|p{2cm}|p{4cm}}
    \toprule[1.5pt]
    & Naturalness & Semantic Faithfulness \\
    \midrule
    AMP\citealp{peng2021amp}-baseline & 3.3 & - \\
    UniHSI-PartNet\citealp{mo2019partnet} & 4.2 & 4.2 \\
    UniHSI-ScanNet\citealp{dai2017scannet} & 3.9 & 4.1 \\
    \bottomrule[1.5pt]
    \end{tabular}}
  \end{center}
\end{table*}
\newpage

\begin{table*}[h!]
  \footnotesize
  \begin{center}
  \caption{\textbf{Exemplification of the LLM Planner through Detailed Prompting.} This caption provides a comprehensive illustration of the input and output of the LLM Planner.}
  \vspace{6pt}
  \label{tab:detailed_prompt_example}
  \scalebox{0.90}{\tablestyle{8pt}{1.0}
    \begin{tabular}{p{420pt}}
    \toprule[1.5pt]
    Input \\
    \midrule
Instruction: I want to play video games for a while, then go to sleep.\\
Background Information:\\
 $[$start of background Information$]$\\
 The room has OBJECTS: $[$bed, chair, table, laptop$]$.\\
The $[$OBJECT: laptop$]$ is upon the $[$OBJECT: table$]$. The $[$OBJECT: table$]$ is in front of the $[$OBJECT: chair$]$. The $[$OBJECT: bed$]$ is several meters away from $[$OBJECT: table$]$. The human is several meters away from these objects.\\
 The $[$OBJECT: bed$]$ has PARTS: $[$pillow, mattress$]$. The $[$OBJECT: chair$]$ has PARTS: $[$back\_soft\_surface, seat\_surface, left\_armrest\_hard\_surface, right\_armrest\_hard\_surface$]$. The $[$OBJECT: table$]$ has PARTS: $[$board$]$. The $[$OBJECT: laptop$]$ has PARTS: $[$screen, keyboard$]$.
 The human has JOINTS: $[$pelvis, left hip, left knee, left foot, right hip, right knee, right foot, torso, head, left shoulder, left elbow, left hand, right shoulder, right elbow, right hand$]$.\\
 $[$end of background Information$]$\\
Given the instruction and background information, generate 1 task plan according to the following rules and examples.\\
 $[$start of rules$]$\\
 1. Each task plan should be composite into detailed steps. If the human is not close to the target object, the first step should be to get close to the object.\\
2. Each step should contain meaningful joint-part pairs.\\
3. Each joint-part pair should be formatted into \{OBJECT, PART, JOINT, Contact type, Contact Direcion\}. Or if the step is getting close to an object, the step should be formatted into \{none, none, none, none, relative direction of the target object\}. JOINT should replace JOINT in the format in the background information. Important: PART in the format should only be replaced by PART or JOINT in the background information. The choices of Contact type include $[$contact, not contact$]$. The choices of Contact Direction include $[$front, back, left, right, up, down, none$]$. \\
4. Be plausible. Do not generate uncommon interactions.\\
5. Only interact with still objects. Do not move objects.\\
$[$end of rules$]$\\
$[$start of an example$]$\\
Instruction: Generate a 6-step interaction with the object.\\
$[$start of background Information$]$\\
The room has a $[$OBJECT: chair$]$.\\
The $[$OBJECT: chair$]$ has PARTS: $[$back\_soft\_surface, armrest\_hard\_surface, armrest\_hard\_surface, seat\_surface$]$.\\
The human has JOINTS: $[$pelvis, left hip, left knee, left foot, right hip, right knee, right foot, torso, head, left shoulder, left elbow, left hand, right shoulder, right elbow, right hand$]$.\\
The human is several meters away from the $[$OBJECT: chair$]$.\\
$[$end of background Information$]$\\
Step 1: Get close to the chair.\\
Pair 1: \{chair, none, none, none, front\}\\
Step 2: Sit on the chair.\\
Pair 1: \{chair, seat\_surface, pelvis, contact, up\}\\
Step 3: Rest the right hand on the right arm of the chair.\\
Pair 1: \{chair, seat\_surfac, pelvis, contact, up\}\\
Pair 2: \{chair, armrest\_hard\_surface, right hand, contact, up\}\\
Step 4: Place the right foot on the left knee.\\
Pair 1: \{chair, seat\_surface, pelvis, contact, up\}\\
Pair 2: \{chair, left knee, right foot, contact, up\}\\
Step 5: Stand up.\\
Pair 1: \{chair, none, none, none, front\}\\
Step 6: Sit on the chair.\\
Pair 1: \{chair, seat\_surface, pelvis, contact, up\}\\
$[$end of an example$]$\\
\midrule
 Output \\
\midrule
Step 1: Get close to the chair.\\
Pair 1: \{chair, none, none, none, front\}\\
Step 2: Sit on the chair.\\
Pair 1: \{chair, seat\_surface, pelvis, contact, up\}\\
Step 3: Use the laptop to play video games.\\
Pair 1: \{chair, seat\_surface, pelvis, contact, up\}\\
Pair 2: \{laptop, keyboard, left hand, contact, up\}\\
Pair 3: \{laptop, keyboard, right hand, contact, up\}\\
Step 4: Stand up from the chair.\\
Pair 1: \{chair, none, none, none, front\}\\
Step 5: Get close to the bed.\\
Pair 1: \{bed, none, none, none, front\}\\
Step 6: Sit on the bed.\\
Pair 1: \{bed, mattress, pelvis, contact, up\}\\
Step 7: Lie down on the bed.\\
Pair 1: \{bed, mattress, pelvis, contact, up\}\\
Pair 2: \{bed, mattress, torso, contact, up\}\\
Pair 3: \{bed, pillow, head, contact, up\}\\

    \bottomrule[1.5pt]
    \end{tabular}}
  \end{center}
\end{table*}

\begin{table*}[h!]
  \footnotesize
  \begin{center}
  \caption{\textbf{Illustration of Simple-Level Interaction Plans in ScenePlan.} Simple-level interaction plans encompass interactions within three steps and involve a single object.}
  \vspace{6pt}
  \label{tab:simple_example} 
  \scalebox{0.90}{\tablestyle{8pt}{1.0}
    \begin{tabular}{p{420pt}}
    \toprule[1.5pt]
\{\\
\qquad ``0000": \\
\qquad\{\\
\qquad\qquad ``obj": \\
\qquad\qquad\{\\
\qquad\qquad\qquad\qquad ``000": \\
\qquad\qquad\qquad\qquad\{\\
\qquad\qquad\qquad\qquad\qquad ``id": ``12747", \\
\qquad\qquad\qquad\qquad\qquad ``name": ``bed", \\
\qquad\qquad\qquad\qquad\qquad ``rotate": $[$$[$1.5707963267948966, 0, 0$]$, $[$0, 0, -1.5707963267948966$]$$]$, \\ 
\qquad\qquad\qquad\qquad\qquad ``scale": 2.5,\\
\qquad\qquad\qquad\qquad\qquad"transfer": $[$0,-2,0$]$,\\
\qquad\qquad\qquad\qquad \}\\
\qquad\qquad\}, \\
\qquad\qquad ``chain\_of\_contacts": $[$$[$$[$``bed000", ``none", ``none", ``none", ``front"$]$$]$, \\
\qquad\qquad\qquad\qquad\qquad\qquad\qquad $[$$[$``bed000", ``mattress25", ``pelvis", ``contact", ``up"$]$,\\
\qquad\qquad\qquad\qquad\qquad\qquad\qquad\qquad\qquad $[$``bed000", ``mattress25", ``head", ``not contact", ``up"$]$$]$, \\
\qquad\qquad\qquad\qquad\qquad\qquad\qquad $[$$[$``bed000", ``mattress25", ``pelvis", ``contact", ``up"$]$, \\
\qquad\qquad\qquad\qquad\qquad\qquad\qquad\qquad\qquad$[$``bed000", ``mattress25", ``left\_foot", ``contact", ``up"$]$, \\
\qquad\qquad\qquad\qquad\qquad\qquad\qquad\qquad\qquad$[$``bed000", ``mattress25", ``right\_foot", ``contact", ``up"$]$, \\
\qquad\qquad\qquad\qquad\qquad\qquad\qquad\qquad\qquad$[$``bed000", ``mattress25", ``head", ``contact", ``up"$]$$]$$]$\\
\qquad\}\\
\}\\
    \bottomrule[1.5pt]
    \end{tabular}}
  \end{center}
\end{table*}

\begin{table*}[h!]
  \footnotesize
  \begin{center}
  \caption{\textbf{Exemplar of Medium-Level Interaction Plans in ScenePlan.} Medium-level interaction plans encompass interactions exceeding three steps and involving a single object.}
  \vspace{6pt}
  \label{tab:mid_example} 
  \scalebox{0.90}{\tablestyle{8pt}{1.0}
    \begin{tabular}{p{420pt}}
    \toprule[1.5pt]
\{\\
\qquad``0000": \\
\qquad\{\\
\qquad\qquad``obj": \{\\
\qquad\qquad\qquad\qquad``000":\{\\
\qquad\qquad\qquad\qquad\qquad``id": ``45005",\\
\qquad\qquad\qquad\qquad\qquad``name": ``chair",\\
\qquad\qquad\qquad\qquad\qquad``rotate": $[$$[$1.5707963267948966, 0, 0$]$, $[$0, 0, -1.5707963267948966$]$$]$,\\
\qquad\qquad\qquad\qquad\qquad``scale": 1.5,\\
\qquad\qquad\qquad\qquad\qquad``transfer": $[$0,-2,0$]$,\\
\qquad\qquad\qquad\qquad\qquad\}\\
\qquad\qquad\qquad\qquad\},\\
\qquad\qquad``chain\_of\_contacts": $[$$[$$[$``chair000", ``none", ``none", ``none", ``front"$]$$]$,\\
\qquad\qquad\qquad\qquad\qquad\qquad\qquad$[$$[$``chair000", ``seat\_soft\_surface42", ``pelvis", ``contact", ``up"$]$$]$,\\
\qquad\qquad\qquad\qquad\qquad\qquad\qquad$[$$[$``chair000", ``seat\_soft\_surface42", ``pelvis", ``contact", ``up"$]$,\\
\qquad\qquad\qquad\qquad\qquad\qquad\qquad$[$``chair000", ``back\_soft\_surface47", ``torso", ``contact", ``none"$]$$]$,\\
\qquad\qquad\qquad\qquad\qquad\qquad\qquad$[$$[$``chair000", ``seat\_soft\_surface42", ``pelvis", ``contact", ``up"$]$,\\
\qquad\qquad\qquad\qquad\qquad\qquad\qquad$[$``chair000", ``back\_soft\_surface47", ``torso", ``contact", ``none"$]$$]$,\\
\qquad\qquad\qquad\qquad\qquad\qquad\qquad$[$$[$``chair000", ``seat\_soft\_surface42", ``pelvis", ``contact", ``up"$]$,\\
\qquad\qquad\qquad\qquad\qquad\qquad\qquad$[$``chair000", ``arm\_sofa\_style44", ``left\_hand", ``contact", ``up"$]$,\\
\qquad\qquad\qquad\qquad\qquad\qquad\qquad$[$``chair000", ``arm\_sofa\_style48", ``right\_hand", ``contact", ``up"$]$$]$,\\
\qquad\qquad\qquad\qquad\qquad\qquad\qquad$[$$[$``chair000", ``seat\_soft\_surface42", ``pelvis", ``contact", ``up"$]$,\\
\qquad\qquad\qquad\qquad\qquad\qquad\qquad$[$``chair000", ``arm\_sofa\_style44", ``left\_hand", ``not contact", ``up"$]$,\\
\qquad\qquad\qquad\qquad\qquad\qquad\qquad$[$``chair000", ``arm\_sofa\_style48", ``right\_hand", ``not contact", ``up"$]$$]$,\\
\qquad\qquad\qquad\qquad\qquad\qquad\qquad$[$$[$``chair000", ``seat\_soft\_surface42", ``pelvis", ``contact", ``up"$]$,\\
\qquad\qquad\qquad\qquad\qquad\qquad\qquad$[$``chair000", ``left\_knee", ``right\_foot", ``contact", ``none"$]$$]$,\\
\qquad\qquad\qquad\qquad\qquad\qquad\qquad$[$$[$``chair000", ``seat\_soft\_surface42", ``pelvis", ``contact", ``up"$]$,\\
\qquad\qquad\qquad\qquad\qquad\qquad\qquad$[$``chair000", ``back\_soft\_surface47", ``torso", ``not contact", ``none"$]$$]$,\\
\qquad\qquad\qquad\qquad\qquad\qquad\qquad$[$$[$``chair000", ``none", ``none", ``none", ``front"$]$$]$$]$\}\\
\}\\
    \bottomrule[1.5pt]
    \end{tabular}}
  \end{center}
\end{table*}

\begin{table*}[h!]
  \footnotesize
  \begin{center}
  \caption{\textbf{An example of hard-level interaction plans in ScenePlan.} Hard-level interaction plans involve interactions of more than 3 steps and more than 1 object.}
  \vspace{6pt}
  \label{tab:long_example} 
  \scalebox{0.90}{\tablestyle{8pt}{1.0}
    \begin{tabular}{p{420pt}}
    \toprule[1.5pt]
\{\\
\qquad``0000": \\
\qquad\{\\
\qquad\qquad``obj": \\
\qquad\qquad\{\\
\qquad\qquad\qquad"000": \\
\qquad\qquad\qquad\{\\
\qquad\qquad\qquad\qquad``id": ``37825", \\
\qquad\qquad\qquad\qquad``name": ``chair",\\
\qquad\qquad\qquad\qquad``rotate": $[$$[$1.5707963267948966, 0, 0$]$, $[$0, 0, -1.5707963267948966$]$$]$,\\
\qquad\qquad\qquad\qquad``scale": 1.5,\\
\qquad\qquad\qquad\qquad``transfer": $[$0,-2,0$]$\\
\qquad\qquad\qquad\},\\
\qquad\qquad\qquad``001": \\
\qquad\qquad\qquad\{\\
\qquad\qquad\qquad\qquad``id": ``21980",\\
\qquad\qquad\qquad\qquad``name": ``table",\\
\qquad\qquad\qquad\qquad``rotate": $[$$[$1.5707963267948966, 0, 0$]$, $[$0, 0, 1.5707963267948966$]$$]$,\\
\qquad\qquad\qquad\qquad``scale": 1.8,\\
\qquad\qquad\qquad\qquad``transfer": $[$1,-2,0$]$\\
\qquad\qquad\qquad\},\\
\qquad\qquad\qquad``002": \\
\qquad\qquad\qquad\{\\
\qquad\qquad\qquad\qquad``id": ``11873",\\
\qquad\qquad\qquad\qquad``name": ``laptop",\\
\qquad\qquad\qquad\qquad``rotate": $[$$[$1.5707963267948966, 0, 0$]$, $[$0, 0, 1.5707963267948966$]$$]$,\\
\qquad\qquad\qquad\qquad``scale": 0.6,\\
\qquad\qquad\qquad\qquad``transfer": $[$0.8,-2,0.65$]$\\
\qquad\qquad\qquad\},\\
\qquad\qquad\qquad``003": \\
\qquad\qquad\qquad\{\\
\qquad\qquad\qquad\qquad``id": ``10873",\\
\qquad\qquad\qquad\qquad``name": ``bed",\\
\qquad\qquad\qquad\qquad``rotate": $[$$[$1.5707963267948966, 0, 0$]$, $[$0, 0, -1.5707963267948966$]$$]$,\\
\qquad\qquad\qquad\qquad``scale": 3,\\
\qquad\qquad\qquad\qquad``transfer": $[$-0.2,-4,0$]$\\
\qquad\qquad\qquad\}\\
\qquad\qquad\}, \\
\qquad\qquad``chain\_of\_contacts": $[$$[$$[$``chair000", ``none", ``none", ``none", ``front"$]$$]$, \\
\qquad\qquad\qquad\qquad\qquad\qquad$[$$[$``chair000", ``seat\_soft\_surface58", ``pelvis", ``contact", ``up"$]$$]$, \\
\qquad\qquad\qquad\qquad\qquad\qquad$[$$[$``chair000", ``seat\_soft\_surface58", ``pelvis", ``contact", ``up"$]$, \\
\qquad\qquad\qquad\qquad\qquad\qquad\qquad$[$``laptop002", ``keyboard15", ``left\_hand", ``contact", ``none"$]$, \\
\qquad\qquad\qquad\qquad\qquad\qquad\qquad$[$``laptop002", ``keyboard15", ``right\_hand", ``contact", ``none"$]$$]$, \\
\qquad\qquad\qquad\qquad\qquad\qquad$[$$[$``chair000", ``none", ``none", ``none", ``front"$]$$]$, \\
\qquad\qquad\qquad\qquad\qquad\qquad$[$$[$``bed003", ``none", ``none", ``none", ``front"$]$$]$, \\
\qquad\qquad\qquad\qquad\qquad\qquad$[$$[$``bed003", ``mattress16", ``pelvis", ``contact", ``up"$]$, \\
\qquad\qquad\qquad\qquad\qquad\qquad\qquad$[$``bed003", ``mattress16", ``head", ``not contact", ``up"$]$$]$, \\
\qquad\qquad\qquad\qquad\qquad\qquad$[$$[$``bed003", ``mattress16", ``pelvis", ``contact", ``up"$]$, \\
\qquad\qquad\qquad\qquad\qquad\qquad\qquad$[$``bed003", ``mattress16", ``left\_foot", ``contact", ``up"$]$, \\
\qquad\qquad\qquad\qquad\qquad\qquad\qquad$[$``bed003", ``mattress16", ``right\_foot", ``contact", ``up"$]$, \\
\qquad\qquad\qquad\qquad\qquad\qquad\qquad$[$``bed003", ``pillow17", ``head", ``contact", ``up"$]$$]$,\\
\qquad\qquad\qquad\qquad\qquad\qquad$[$$[$``bed003", ``mattress16", ``pelvis", ``contact", ``up"$]$, \\
\qquad\qquad\qquad\qquad\qquad\qquad\qquad$[$``bed003", ``mattress16", ``head", ``not contact", ``up"$]$$]$,\\
\qquad\qquad\qquad\qquad\qquad\qquad\qquad$[$$[$``bed003", ``none", ``none", ``none", ``front"$]$$]$$]$\\
\qquad\}\\
\}\\
    \bottomrule[1.5pt]
    \end{tabular}}
  \end{center}
\end{table*}

\begin{table*}[h!]
  \footnotesize
  \begin{center}
  \caption{List of Interactions in ScenePlan-1}
  \vspace{6pt}
  \label{tab:sceneplan_list_1} 
  \scalebox{0.90}{\tablestyle{8pt}{1.0}
    \begin{tabular}{p{180pt}p{180pt}}
    \toprule[1.5pt]
    \textbf{Interaction Type} & \textbf{Contact Formation} \\
    \midrule[1.5pt]
    Get close to xxx & \{xxx, none, none, none, dir\} \\
    \midrule
    Stand up & \{xxx, none, none, none, dir\} \\
    \midrule
    Left hand reaches xxx & \{xxx, part, left\_hand, contact, dir\} \\
    \midrule
    Right hand reaches xxx & \{xxx, part, right\_hand, contact, dir\} \\
    \midrule
    Both hands reaches xxx & \makecell[l]{\{\{xxx, part, left\_hand, contact, dir\}, \\
                                                    \{xxx, part, right\_hand, contact, dir\}\}} \\
    \midrule
    Sit on xxx & \{xxx, seat\_surface, pelvis, contact, up\} \\
    \midrule
    Sit on xxx, left hand on left arm & \makecell[l]{\{\{xxx, seat\_surface, pelvis, contact, up\}, \\
                                                    \{xxx, left\_arm, left\_hand, contact, up\}\}} \\
    \midrule
    Sit on xxx, right hand on right arm & \makecell[l]{\{\{xxx, seat\_surface, pelvis, contact, up\},\\
                                                    \{xxx, right\_arm, right\_hand, contact, up\}\}} \\
    \midrule
    Sit on xxx, hands on arms & \makecell[l]{\{\{xxx, seat\_surface, pelvis, contact, up\}, \\
                                                    \{xxx, left\_arm, left\_hand, contact, none\},\\
                                                    \{xxx, right\_arm, right\_hand, contact, none\}\}} \\
    \midrule
    Sit on xxx, hands away from arms & \makecell[l]{\{\{xxx, seat\_surface, pelvis, contact, up\}, \\
                                                    \{xxx, left\_arm, left\_hand, not contact, none\},\\
                                                    \{xxx, right\_arm, right\_hand, not contact, none\}\}} \\
    \midrule
    Sit on xxx, left elbow on left arm & \makecell[l]{\{\{xxx, seat\_surface, pelvis, contact, up\}, \\
                                                    \{xxx, left\_arm, left\_elbow, contact, up\}\}} \\
    \midrule
    Sit on xxx, right elbow on right arm & \makecell[l]{\{\{xxx, seat\_surface, pelvis, contact, up\},\\
                                                    \{xxx, right\_arm, right\_elbow, contact, up\}\}} \\
    \midrule
    Sit on xxx, elbows on arms & \makecell[l]{\{\{xxx, seat\_surface, pelvis, contact, up\}, \\
                                                    \{xxx, left\_arm, left\_elbow, contact, none\},\\
                                                    \{xxx, right\_arm, right\_elbow, contact, none\}\}} \\
    \midrule
    Sit on xxx, left hand on left knee & \makecell[l]{\{\{xxx, seat\_surface, pelvis, contact, up\}, \\
                                                    \{xxx, left\_knee, left\_hand, contact, up\}\}} \\
    \midrule
    Sit on xxx, right hand on right knee & \makecell[l]{\{\{xxx, seat\_surface, pelvis, contact, up\},\\
                                                    \{xxx, right\_knee, right\_hand, contact, up\}\}} \\
    \midrule
    Sit on xxx, hands on knees & \makecell[l]{\{\{xxx, seat\_surface, pelvis, contact, up\}, \\
                                                    \{xxx, left\_knee, left\_hand, contact, none\},\\
                                                    \{xxx, right\_knee, right\_hand, contact, none\}\}} \\
    \midrule
    Sit on xxx, left hand on stomach & \makecell[l]{\{\{xxx, seat\_surface, pelvis, contact, up\}, \\
                                                    \{xxx, pelvis, left\_hand, contact, none\}\}} \\
    \midrule
    Sit on xxx, right hand on stomach & \makecell[l]{\{\{xxx, seat\_surface, pelvis, contact, up\},\\
                                                    \{xxx, pelvis, right\_hand, contact, none\}\}} \\
    \midrule
    Sit on xxx, hands on stomach & \makecell[l]{\{\{xxx, seat\_surface, pelvis, contact, up\}, \\
                                                    \{xxx, pelvis, left\_hand, contact, none\},\\
                                                    \{xxx, pelvis, right\_hand, contact, none\}\}} \\
    \midrule
    Sit on xxx, left foot on right knee & \makecell[l]{\{\{xxx, seat\_surface, pelvis, contact, up\}, \\
                                                    \{xxx, right\_knee, left\_foot, contact, none\}\}} \\
    \midrule
    Sit on xxx, right foot on left knee & \makecell[l]{\{\{xxx, seat\_surface, pelvis, contact, up\}, \\
                                                    \{xxx, left\_knee, right\_foot, contact, none\}\}} \\
    \midrule
    Sit on xxx, lean forward & \makecell[l]{\{\{xxx, seat\_surface, pelvis, contact, up\}, \\
                                                    \{xxx, back\_surface, torso, not contact, none\}\}} \\
    \midrule
    Sit on xxx, lean backward & \makecell[l]{\{\{xxx, seat\_surface, pelvis, contact, up\}, \\
                                                    \{xxx, back\_surface, torso, contact, none\}\}} \\
    \bottomrule[1.5pt]
    \end{tabular}}
  \end{center}
\end{table*}

\begin{table*}[h!]
  \footnotesize
  \begin{center}
  \caption{List of Interactions in ScenePlan-2}
  \vspace{6pt}
  \label{tab:sceneplan_list_2}
  \scalebox{0.90}{\tablestyle{8pt}{1.0}
    \begin{tabular}{p{180pt}p{180pt}}
    \toprule[1.5pt]
    \textbf{Interaction Type} & \textbf{Contact Formation} \\
    \midrule[1.5pt]
    Lie on xxx & \makecell[l]{\{\{xxx, mattress, pelvis, contact, up\}, \\
                                \{xxx, pillow, head, contact, up\}\}} \\    
    \midrule
    Lie on xxx, left knee up & \makecell[l]{\{\{xxx, mattress, pelvis, contact, up\}, \\
                                \{xxx, pillow, head, contact, up\\
                                \{xxx, mattress, left\_knee, not contact, none\}\}} \\
    \midrule
    Lie on xxx, right knee up & \makecell[l]{\{\{xxx, mattress, pelvis, contact, up\}, \\
                                \{xxx, pillow, head, contact, up\},\\
                                \{xxx, mattress, right\_knee, not contact, none\}\}} \\
    \midrule
    Lie on xxx, knees up & \makecell[l]{\{\{xxx, mattress, pelvis, contact, up\}, \\
                                \{xxx, pillow, head, contact, up\},\\
                                \{xxx, mattress, left\_knee, not contact, none\},\\
                                \{xxx, mattress, right\_knee, not contact, none\}\}} \\
    \midrule
    Lie on xxx, left hand on pillow & \makecell[l]{\{\{xxx, mattress, pelvis, contact, up\}, \\
                                \{xxx, pillow, head, contact, up\},\\
                                \{xxx, pillow, left\_hand, contact, none\}\}} \\
    \midrule
    Lie on xxx, right hand on pillow & \makecell[l]{\{\{xxx, mattress, pelvis, contact, up\}, \\
                                \{xxx, pillow, head, contact, up\},\\
                                \{xxx, pillow, right\_hand, contact, none\}\}} \\
    \midrule
    Lie on xxx, hands on pillow & \makecell[l]{\{\{xxx, mattress, pelvis, contact, up\}, \\
                                \{xxx, pillow, head, contact, up\},\\
                                \{xxx, pillow, left\_hand, contact, none\},\\
                                \{xxx, pillow, right\_hand, contact, none\}\}} \\
    \midrule
    Lie on xxx, on left side & \makecell[l]{\{\{xxx, mattress, pelvis, contact, up\}, \\
                                \{xxx, pillow, head, contact, up\},\\
                                \{xxx, mattress, right\_shoulder, not contact, none\}\}} \\
    \midrule
    Lie on xxx, on right side & \makecell[l]{\{\{xxx, mattress, pelvis, contact, up\}, \\
                                \{xxx, pillow, head, contact, up\},\\
                                \{xxx, mattress, left\_shoulder, not contact, none\}\}} \\
    \midrule
    Lie on xxx, left foot on right knee & \makecell[l]{\{\{xxx, mattress, pelvis, contact, up\}, \\
                                \{xxx, pillow, head, contact, up\},\\
                                \{xxx, right\_knee, left\_foot, contact, up\}\}} \\
    \midrule
    Lie on xxx, right foot on left knee & \makecell[l]{\{\{xxx, mattress, pelvis, contact, up\}, \\
                                \{xxx, pillow, head, contact, up\},\\
                                \{xxx, left\_knee, right\_foot, contact, up\}\}} \\
    \midrule
    Lie on xxx, head up & \makecell[l]{\{\{xxx, mattress, pelvis, contact, up\}, \\
                                \{xxx, pillow, head, not contact, none\}\}}\\
    \bottomrule[1.5pt]

    \end{tabular}}
  \end{center}
\end{table*}


\bibliography{mainbib}
\bibliographystyle{iclr2024_conference}

\end{document}